\documentclass[sigconf]{acmart}
\usepackage[utf8]{inputenc}

\setcopyright{none}

\usepackage{hyperref}
\usepackage{subcaption}
\usepackage{xspace}
\usepackage{booktabs}
\usepackage{multirow}
\usepackage{makecell}
\usepackage{subcaption}
\usepackage{float}
\newcommand{\approach}{\textsc{Ovid}\xspace}

\renewcommand\footnotetextcopyrightpermission[1]{}

\title{\approach: A Machine Learning Approach for Automated Vandalism Detection in OpenStreetMap}

\newcommand\blfootnote[1]{%
  \begingroup
  \renewcommand\thefootnote{}\footnote{#1}%
  \addtocounter{footnote}{-1}%
  \endgroup
}

\author{Nicolas Tempelmeier}
\affiliation{%
  \institution{L3S Research Center\\Leibniz University Hannover}
  \city{Hannover}
  \country{Germany}
} \email{tempelmeier@L3S.de}

\author{Elena Demidova}
\affiliation{%
  \institution{Data Science \& Intelligent Systems (DSIS)\\ University of Bonn}
  \city{Bonn}
  \country{Germany}}
\email{elena.demidova@cs.uni-bonn.de}


\newcommand{\random}{\textsc{Random}\xspace}
\newcommand{\patrol}{\textsc{OSMPatrol}\xspace}
\newcommand{\watchman}{\textsc{OSMWatchman}\xspace}
\newcommand{\WDVD}{\textsc{WDVD}\xspace}
\newcommand{\Glove}{\textsc{GloVe+CNN}\xspace}


\begin{document}

\begin{abstract}
OpenStreetMap is a unique source of openly available worldwide map data, increasingly adopted in real-world applications. 
Vandalism detection in OpenStreetMap is critical and remarkably challenging due to the large scale of the dataset, the sheer number of contributors, various vandalism forms, and the lack of annotated data to train machine learning algorithms.
This paper presents \approach{} - a novel machine learning method for vandalism detection in OpenStreetMap. 
\approach{} relies on a neural network architecture that adopts a multi-head attention mechanism to effectively summarize information indicating vandalism from OpenStreetMap changesets.
To facilitate automated vandalism detection, we introduce a set of original features that capture changeset, user, and edit information.
Our evaluation results on real-world vandalism data demonstrate that the proposed \approach{} method outperforms the baselines by 4.7 percentage points in F1 score.
\end{abstract}

\maketitle
\pagestyle{plain}
\blfootnote{\textcopyright Nicolas Tempelmeier, Elena Demidova 2021. This is the author's version of the work. It is posted here for your
personal use. Not for redistribution. The definitive version was published in the proceedings of the 29th ACM SIGSPATIAL International Conference on Advances in Geographic Information Systems \url{https://doi.org/10.1145/3474717.3484204}.\\}

\section{Introduction}
\label{sec:intro}

\begin{figure*}
    \centering
    \begin{subfigure}[t]{0.42\textwidth}
        \includegraphics[width=\textwidth]{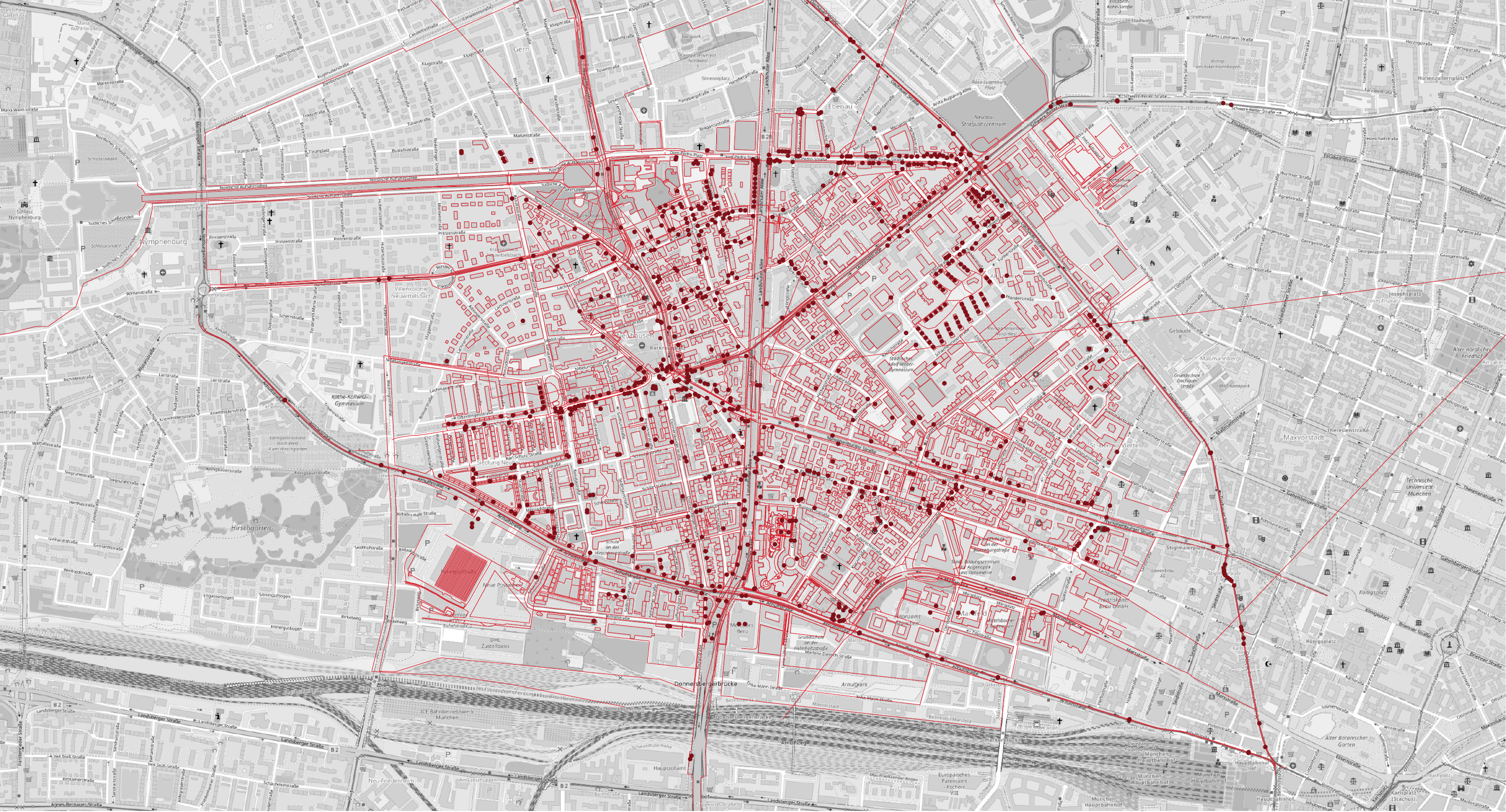}
        \caption{Deletion of large fractions of Munich}
    \end{subfigure}
    \hspace{0.1\textwidth}
    \begin{subfigure}[t]{0.42\textwidth}
        \includegraphics[width=\textwidth]{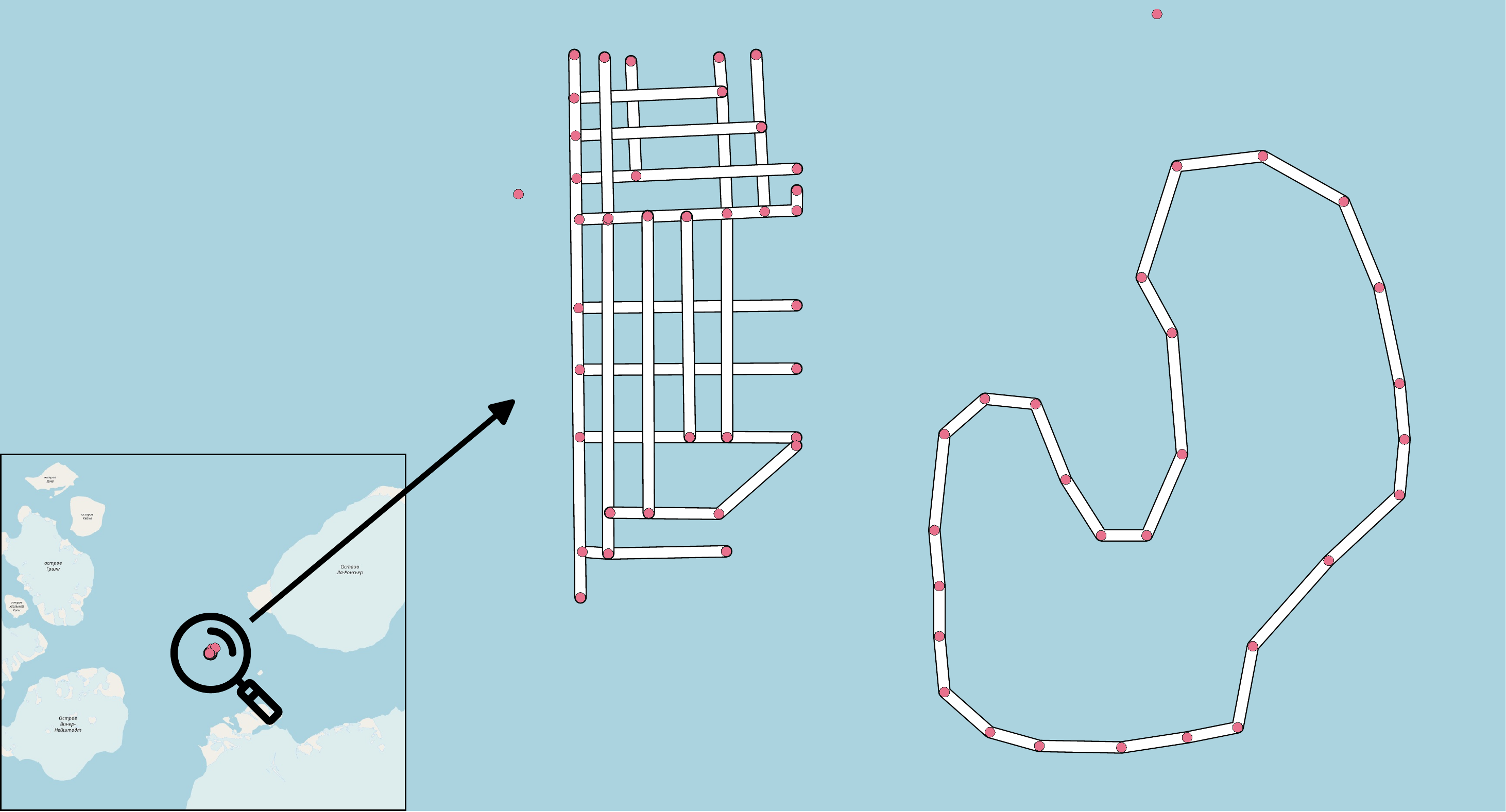}
        \caption{A fake town within the ocean}
    \end{subfigure}
    \caption{Real-world examples of different vandalism forms in OSM. Map data: \textcopyright OpenStreetMap contributors, ODbL.}
    \label{fig:intro_example}
\end{figure*}

%

OpenStreetMap (OSM) has evolved as a critical source of openly available volunteered geographic information \cite{JokarArsanjani2015}.
The amount of geospatial information in OpenStreetMap is continuously growing. For instance, the number of nodes captured by OSM increased from $5.9 \cdot 10^9$ in March 2020 to $6.7 \cdot 10^9$ in March 2021.
With the OSM growth, quality assurance becomes essential but also increasingly challenging.
OSM is a collaborative online project aiming to create a free and editable map.
Today, OSM provides a data basis for various real-world applications, including navigation systems and geographic information systems (GIS).
Detecting and removing vandalism cases is essential to preserve the credibility and trust in OSM data.

In OSM, contributors voluntarily provide geographic information and can add, modify or delete any objects captured by OSM. This openness makes the data particularly susceptible to vandalism.
Recently, the problem of vandalism detection in OSM has attracted interest of researchers \cite{ijgi9090504} and OSM contributors\footnote{\url{https://wiki.openstreetmap.org/wiki/Vandalism}}.
For example, the OSM community identified several cases of vandalism in the context of the location-based mobile game Pokémon Go \cite{ijgi9040197}, in which users added wrong information to the map to gain an advantage in the game.
The problem of vandalism detection in OSM is particularly challenging due to the large scale of the dataset, the high number of contributors (over 7.6 million in June 2021), the variety of forms vandalism can take, and the lack of annotated data to 
train machine learning algorithms.

Figure \ref{fig:intro_example} presents two real-world vandalism examples.
The vandalism forms in OSM include the arbitrary deletion of map regions and creating non-existing cities in the middle of an ocean.
Vandalism detection methods need to consider various aspects such as the geographic context, typical user behaviour, and content semantics to identify potentially malicious edits effectively.
The diversity of vandalism appearances and relevant features constitutes a significant challenge for automated vandalism detection.

Whereas the existing literature has considered OSM vandalism previously, only a few automated approaches for vandalism detection in OSM exist.
An early approach proposed in \cite{ijgi1030315} adopts a rule-based method to identify suspicious edits. This approach is subject to numerous manually tuned thresholds. 
In \cite{ijgi9090504}, the authors proposed a random forest-based method that detects vandalized buildings. This approach is limited to the building domain and does not capture other various OSM vandalism forms. 
Furthermore, due to the shortage of benchmark datasets with real vandalism examples, existing studies typically utilize synthetic data and lack evaluation in real-world settings.

In this paper, we present the \approach{} (\underline{O}penStreetMap \underline{V}andal\underline{i}sm \underline{D}etection) model - a novel supervised machine learning approach to detect a variety of vandalism forms in OSM effectively.
We propose a neural network architecture that adopts multi-head attention to select the most relevant edits within an individual changeset, i.e., a set of edits performed by a user within one session.
Furthermore, we propose an original feature set that captures different aspects of OSM vandalism, such as user experience and contribution content.
We train and evaluate \approach on real-world vandalism occurrences in OSM manually identified by the OSM community.
The main contributions of this paper are as follows: (i) We present \approach{} -- a novel machine learning method for vandalism detection in OpenStreetMap. (ii) We conduct an evaluation on the extracted real-world vandalism dataset and demonstrate the effectiveness of the proposed \approach{} method, outperforming the baselines by 4.7 percentage points in F1 score.

\section{Background \& Problem Definition}
\label{sec:problem}

OpenStreetMap categorizes the geographic objects it contains into three types. 
\emph{Nodes} represent geographic points (e.g., mountain peaks) with the position specified by latitude and longitude.
\emph{Ways} represent lines (e.g., roads) composed of a sequence of nodes. 
\emph{Relations} are composed of nodes and ways and describe more complex objects, e.g., national borders.
Relations can also include sub-relations.
An OSM object may exhibit an arbitrary number of \emph{tags}, i.e., key-value pairs, that describe the semantics of the object. 
For instance, the tag \texttt{place=city} indicates that an OSM object annotated with this tag represents a city.

\begin{figure*}
    \centering
    \includegraphics[width=0.8\textwidth]{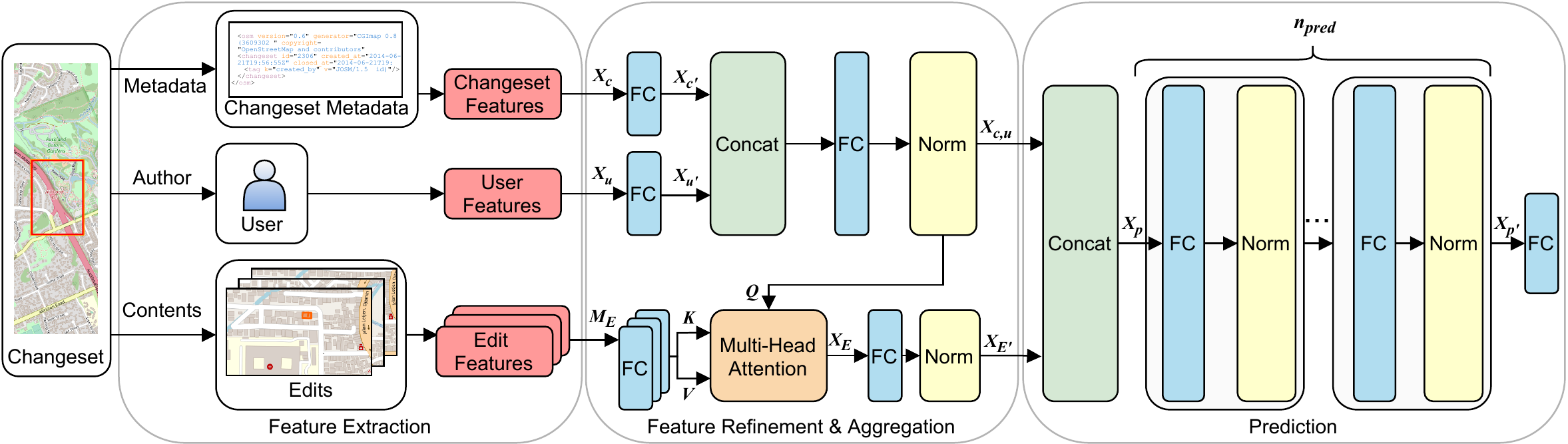}
    \caption{\approach{} model architecture. The changeset and user feature refinement part, as well as the prediction layers, are composed of fully connected (FC), normalization (norm), and concatenation (concat) layers. Multi-head attention layers aggregate the features from multiple edits in a changeset into a single feature vector. Map images: \textcopyright OpenStreetMap contributors, ODbL}
    \label{fig:model_architecture}
\end{figure*}

Formally, an \emph{OSM object} is defined as $o = \langle id, type, loc, tags, ver \rangle$, where:
$id$ is an object identifier.
$type \in \{\texttt{Node}, ~\texttt{Way}, ~\texttt{Relation} \}$ indicates the object type.
$loc$ indicates the geographic location of the object. The location can either be a point (Node), a line (Way) or a set of points and lines (Relation).
$tags$ is a set of tags describing object characteristics. Each tag $\langle k,v \rangle \in tags$ is represented as a key-value pair with the key $k$ and a value $v$.
$ver$ is the version number of the object. The version number corresponds to the number of revisions of the object $o$.
An OSM object $o$ can be distinguished by its identifier $o.id$ together with its type $o.type$.

OSM allows for updates in the form of edits of individual OSM objects. 
An edit can either \emph{create} new objects or \emph{modify} or \emph{delete} existing objects.
More formally, an \emph{edit} is defined as $e=\langle o, op, ver, t\rangle $, where: 
$o$ is an OpenStreetMap object, 
$op \in \{\texttt{create}, \texttt{modify}, \texttt{delete}\}$ is the operation performed on the object $o$, 
$ver$ is the new version number ($o.ver+1$) after the edit is performed, 
and $t$ is the time when the edit took place.

Edits are submitted to OSM in the form of \emph{changesets}. 
Changesets bundle multiple edits created by a single user during a short time period. 
A \emph{changeset} is defined as $c=\langle E, t, u, co \rangle$, where:
$E$ is a set of OSM edits that belong to the changeset,
$t$ is the changeset commit time,
$u$ is the user who committed the changeset,
and $co$ is a comment describing the changeset contents.
We denote the set of all changesets by $C$.

In this paper, we target the problem of identifying vandalism changesets in OpenStreetMap, i.e., the changesets that contain wrong or prohibited (e.g., discriminating or offensive) content.
Formally, \emph{vandalism detection} is the task of identifying changesets that constitute vandalism by either deleting correct information or adding wrong or prohibited information. We aim to learn a function $\hat{y}:C \mapsto \{\text{True}, \text{False}\}$ that assigns vandalism labels to changesets.
In OSM, examples of vandalism include the deletion of existing towns, creating non-existing roads, adding advertisements, or replacing object names with offensive terms. Some vandalism examples are illustrated in Figure \ref{fig:intro_example}.

\section{The \approach{} Model}
\label{sec:approach}

This section presents the \approach{} (\underline{O}penStreetMap \underline{V}andal\underline{i}sm \underline{D}etection) model.
\approach{} is a supervised binary classification model that discriminates between regular and vandalism OSM changesets. 
The model consists of a supervised artificial neural network, including three main components: Feature Extraction, Feature Refinement \& Aggregation, and Prediction.
Figure \ref{fig:model_architecture} provides an overview of the \approach{} model architecture.
We adopt features in three categories. 
First, changeset features capture meta-information of the changesets, e.g., the editor software.
Second, user features provide information regarding previous editing activities of the changeset author, e.g., the number of prior contributions.
Third, edit features encode information describing the individual changes within the changeset, e.g., if an object was added, modified, or deleted.
The following section describes present \approach{}'s components in more detail.

\subsection{Feature Extraction}
\label{sec:feature-extraction}

This section describes the extraction of the changeset, user, and edit features.
\subsubsection{Changeset Features}
The changeset features provide information regarding the changeset metadata.
We consider the following changeset features:
Features capturing the general \emph{changeset content}, such as, no. creates, no. modifications, no. deletes, and no. edits.
Features capturing the \emph{geographic extent}, such as min/max latitude/longitude and bounding box size.
Features capturing the \emph{creation process} such as editor application, comment length, and has imagery used.
We denote the changeset feature vector by $X_{c}$.

\subsubsection{User Features}
We utilize user features to capture the previous activity of the changeset author $c.u$, as a more experienced user may be more trustworthy than a new user.
We consider the following user features:
Features capturing the \emph{past contributions}, such as no. past creates, no. past modifications, no. past deletes, and no. contributions.
Features capturing \emph{user experience}, such as account creation date, and no. active weeks.
Features capturing the \emph{use of OSM conventions}, such as no. top-12 keys used.

Given a changeset $c$ and its author $c.u$, we denote the user feature vector by $X_u$.

\subsubsection{Edit Features}
The edit features capture information regarding the individual edits contained in a changeset. 
We extract the following features from each edit $e$: 
Features capturing the \emph{object characteristics}, such as object type, object version number,  no. previous authors, no. tags, and no. valid tags.
Features capturing the \emph{difference} to the previous OSM version, such as edit operation, time to the previous version, no. previous valid tags, and name changed.

For each edit $e \in c.E$, we concatenate the features into an individual edit feature vector $X_{e}$.


\subsection{Feature Refinement \& Aggregation}
\label{sec:feature-refinement}

In this step, we first refine the changeset and user features and then aggregate the edit features to obtain a single feature vector.

We refine the changeset and user feature vectors by passing them to the fully connected layers with ReLU activation $X_{c'} = FC(X_{c})$ and  $X_{u'} = FC(X_u)$.
We concatenate the changeset and user features and apply a fully connected layer with normalization:
$
    X_{c,u} = norm(FC([X_{c'}, X_{u'}])).
$

We aim at selecting the edits most relevant to identify vandalism in the corresponding changeset.
We combine the feature vectors $X_e$ for each $e \in c.E$ into the edit feature matrix $M_e$.
We apply the same fully connected layer to each edit $M_{e'} = FC(M_e)$ to obtain the refined edit features $M_{e'}$.
To aggregate the features of the individual edits into a single feature vector, we adopt the multi-head attention mechanism proposed by \cite{NIPS2017_3f5ee243}. 
As we aim at selecting the edits most relevant to identify vandalism in the corresponding changeset,
we represent the refined changeset and user features as the query in the attention model
and the refined edit features as keys and values:
$Q = X_{c,u}$, 
$K = M_{e'}$, and
$V = M_{e'}$.
We compute an aggregated edit feature vector using multi-head attention: $X_E = \textit{Multi-Head}(X_{c,u}, M_{e'}, M_{e'})$ as defined in \cite{NIPS2017_3f5ee243}.
We refine the edit feature vector using a fully connected layer with the layer normalization $X_{E'} = norm(FC(X_E))$.
Finally, we introduce an upper threshold $th_{e,max}$ for the maximum number of edits within a changeset. 
If the number of edits exceeds $th_{e,max}$, we set $X_{E'}=0$ and rely on the user and changeset features.

\subsection{Prediction}
\label{sec:prediction-layers}

We facilitate the detection of vandalism changesets by combining the intermediate vectors into a single feature vector $X_p = [X_{c,u}, X_{E'}]$.
We repeat fully connected layers with layer normalization $n_{pred}$ times and use a final fully connected layer with a single output dimension and sigmoid activation function to make predictions.
$
    X_{p'} = norm(FC(X_p))^{n_{pred}},
$
$
    \hat{y} = sigmoid(X_{p'} W_{p'}\ + b_{p'}),
$
with the weight matrix $W_{p'}$ and the bias vector $b_{p'}$.

\section{Evaluation Setup}
\label{sec:setup}
This section describes the dataset, baselines, and metrics.

\subsection{Dataset}
\label{sec:datasets}
We create a ground truth dataset by considering the changesets reverting vandalism in the OpenStreetMap history from 2014 to 2019.
The extraction process consists of the following steps:
First, we extract the revert changesets that fix vandalism changesets.
We only consider changesets that mention ``vandalism'' in their comments.
Second, we determine the vandalism changesets corrected by the revert.
If a revert changeset explicitly mentions a specific changeset, we consider the mentioned changeset as vandalism.
Otherwise, we review the objects that are the subject of the revert.
If the revert deletes an object and only one user contributed to the object, we consider changesets contributing to this object to be vandalism.

To create negative examples (i.e., changesets that do not represent vandalism), we remove the identified vandalism changesets and the reverts from the OSM changeset history.
We randomly sample the same number of changesets as the vandalism changesets from the reduced changeset history to create negative examples and obtain a balanced dataset.
We split the dataset into training (70\%), validation (10\%), and test (20\%) sets.
To avoid bias towards individual OSM users, we ensure that the training, validation, and test sets are disjunct concerning OSM users.

\subsection{Baselines}
\label{sec:baseline}
We compare our model with the following baselines:\\
\textbf{\random}. This baseline chooses the vandalism label at random.\\
\textbf{\patrol}. This model is an early approach to detect vandalism in OSM \cite{ijgi1030315}. 
OSMPatrol is a rule-based system aiming to identify vandalism at the level of OSM edits.
In our settings, we utilize this baseline to label changesets.
We consider a changeset as vandalism if at least one edit in this changeset is vandalism.\\
\textbf{\watchman}. This model was recently proposed to detect vandalism on buildings in OSM \cite{ijgi9090504}. 
\watchman uses a random forest classifier that utilizes content features, context features, and user features.\\
\textbf{\WDVD}. The Wikidata Vandalism Model was proposed to detect vandalism in  Wikidata \cite{10.1145/2983323.2983740}. 
This baseline uses a random forest classifier with text-based features to detect vandalism.\\
\textbf{\Glove}. OSM community members developed this baseline to detect suspicious changesets \cite{glovecnn}.
This baseline uses GloVe embeddings \cite{pennington2014glove} to represent OSM changesets as input for a CNN.

\begin{table}

\caption{Vandalism detection performance with respect to precision, recall, F1 score and accuracy [\%]. Best scores are marked bold.}
\begin{tabular}{l@{\quad}cccc}
\toprule
Approach &  Precision & Recall & F1 & Accuracy  \\  
\midrule

\random  & 49.89 & 50.27  & 50.08 & 49.92\\
\patrol   & 53.94 &  \textbf{96.29} & 69.15 & 57.06 \\
\watchman   & 77.60 & 70.74  & 74.01 & 75.18\\
\WDVD   &  \textbf{81.52} & 64.79  & 72.20 & 75.07\\
\Glove   & 81.46 & 72.93 & 76.96 & 78.18 \\

\midrule

\textsc{\approach} & 80.35 & 83.02  & \textbf{81.66} & \textbf{81.37}\\
     \bottomrule
\end{tabular}
\label{tab:results}

\end{table}

\subsection{Metrics}
\label{sec:metrics}
To evaluate the performance of the different vandalism detection approaches, we compute the following metrics:\\
\textbf{Precision.} The fraction of correctly classified vandalism instances among all instances classified as vandalism.\\
\textbf{Recall.} The fraction of correctly classified vandalism instances among all vandalism instances.\\
\textbf{F1 score.} The harmonic mean of recall and precision.\\
\textbf{Accuracy.} The fraction of correctly classified instances among all instances.

\newcommand{\approachsc}{\approach}

\section{Evaluation Results}
\label{sec:evaluation}

The evaluation aims to assess the effectiveness of the proposed \approach{} approach for vandalism detection.
Table \ref{tab:results} summarizes the overall vandalism detection performance.

Overall, we observe that in terms of F1 score and accuracy, \approachsc achieves the best performance.
\approachsc achieves 4.7 percent points improvement in F1 score and 3.19 percent points improvement in accuracy compared to the best performing baseline.

The \patrol baseline achieves the best recall.
However, \patrol's precision score is close to 50\%, which corresponds to the performance of the random choices by the na\"ive \random baseline. 
The recall and precision scores reveal that \patrol assigns almost all changesets to the vandalism class, resulting in ultimately low accuracy of 57.06\%.
The low accuracy scores indicate that supervised machine learning models like \approachsc are better suited to detect vandalism than the \patrol system that relies on manually specified rules.

\WDVD achieves the best precision, but only reaches a recall score of 64.79\%.
\WDVD mainly relies on user features. 
The high precision score indicates that user features can effectively identify a fraction of the malicious changesets.
However, the low recall score indicates that user features are insufficient to capture the diverse vandalism forms.
In contrast, \approach that considers user, changeset, and edit features, achieves 83.02\% recall.
The \watchman baseline achieves a moderate performance considering all metrics.
\Glove achieves the best baseline performance in F1 score and accuracy by using information from the changeset and the edits, but does not consider user information.
\approachsc that combines user and content features achieves a high recall (83.02\%) while maintaining a comparably high precision (80.35\%).

\section{Conclusion}
\label{sec:conclusion}

In this paper, we proposed the \approach (\underline{O}penStreetMap \underline{V}andal\underline{i}sm \underline{D}etection) model, a novel supervised machine learning approach for vandalism detection in OpenStreetMap. 
\approach relies on original changeset, user, and edit features  to identify vandalism changesets in OSM effectively.
Our experiments on a real-world dataset demonstrate that \approach can effectively detect OSM vandalism.

\subsubsection*{Acknowledgements} This work was partially funded by DFG, The German Research Foundation (``WorldKG'', 424985896), 
the Federal Ministry for Economic Affairs and Energy (BMWi), Germany (``d-E-mand'', 01ME19009B), 
and the European Commission (EU H2020, ``smashHit'', grant-ID 871477).

\bibliographystyle{ACM-Reference-Format}
\bibliography{ref}

\end{document}